\documentclass[letterpaper]{article} % DO NOT CHANGE THIS
\usepackage{aaai24}  % DO NOT CHANGE THIS
\usepackage{times}  % DO NOT CHANGE THIS
\usepackage{helvet}  % DO NOT CHANGE THIS
\usepackage{courier}  % DO NOT CHANGE THIS
\usepackage[hyphens]{url}  % DO NOT CHANGE THIS
\usepackage{graphicx} % DO NOT CHANGE THIS
\usepackage{natbib}  % DO NOT CHANGE THIS AND DO NOT ADD ANY OPTIONS TO IT
\usepackage{caption} % DO NOT CHANGE THIS AND DO NOT ADD ANY OPTIONS TO IT
\usepackage{algorithmic}
\usepackage{booktabs}
\usepackage[ruled,vlined,linesnumbered]{algorithm2e}
\usepackage{xcolor}
\definecolor{codegreen}{rgb}{0.0,0.6,0.0}
\usepackage{lipsum}
\usepackage{diagbox}
\usepackage{amsmath}
\usepackage{xcolor}
\usepackage{subfig}
\usepackage{tabularray}
\usepackage{multirow}

\usepackage{newfloat}
\usepackage{float}
\usepackage{listings}
\DeclareCaptionStyle{ruled}{labelfont=normalfont,labelsep=colon,strut=off} % DO NOT CHANGE THIS
\floatstyle{ruled}
\newfloat{listing}{tb}{lst}{}
\floatname{listing}{Listing}
\newcommand{\system}{SlowTrack\xspace}

\usepackage[misc]{ifsym}
\title{SlowTrack: Increasing the Latency of Camera-based Perception in Autonomous Driving Using Adversarial Examples}
\author{Chen Ma\textsuperscript{\rm 1}\equalcontrib \ \ \ \ Ningfei Wang\textsuperscript{\rm 2}\equalcontrib \ \ \ \ Qi Alfred Chen\textsuperscript{\rm 2} \ \ \ \ Chao Shen\textsuperscript{\rm 1}}
\affiliations{
\textsuperscript{\rm 1}Xi'an Jiaotong University \ \ \ 
\textsuperscript{\rm 2}University of California, Irvine \\
ershang@stu.xjtu.edu.cn \ \ \ ningfei.wang@uci.edu \ \ \ alfchen@uci.edu \ \ \ chaoshen@xjtu.edu.cn}

\usepackage{bibentry}
\begin{document}

\maketitle

\begin{abstract}
In Autonomous Driving (AD), real-time perception is a critical component responsible for detecting surrounding objects to ensure safe driving. While researchers have extensively explored the integrity of AD perception due to its safety and security implications, the aspect of availability (real-time performance) or latency has received limited attention. Existing works on latency-based attack have focused mainly on \textit{object detection}, i.e., a component in camera-based AD perception, overlooking the entire camera-based AD perception, which hinders them to achieve effective system-level effects, such as vehicle crashes. In this paper, we propose \system, a novel framework for generating adversarial attacks to increase the execution time of camera-based AD perception. We propose a novel two-stage attack strategy along with the three new loss function designs. Our evaluation is conducted on four popular camera-based AD perception pipelines, and the results demonstrate that \system significantly outperforms existing latency-based attacks while maintaining comparable imperceptibility levels. Furthermore, we perform the evaluation on Baidu Apollo, an industry-grade full-stack AD system, and LGSVL, a production-grade AD simulator, with two scenarios to compare the system-level effects of \system and existing attacks. Our evaluation results show that the system-level effects can be significantly improved, i.e., the vehicle crash rate of \system is around 95\% on average while existing works only have around 30\%.

\end{abstract}
\section{Introduction}
\label{sec:intro}

\hspace*{10pt} 
Autonomous Driving (AD) vehicles, manufactured by various companies, have become ubiquitous in our daily lives. For instance, numerous Tesla vehicles are equipped with the Autopilot feature~\cite{Kane2021Tesla, Tesla2022autopilot} running in the real world. For these vehicles, camera-based \textit{perception} is pivotal, enabling them to detect real-time environmental objects such as pedestrians to ensure safety. 
Given its significance for safety and security, various prior works~\cite{sp:2021:ningfei:msf-adv, shen2022sok, sato2021dirty, Wang_2023_ICCV} have studied its security, especially on integrity such as making the object vanished or changing the label of the objects to cause traffic rule violations or safety hazards. We refer to these as \textit{system-level effects} throughout this paper.

Nevertheless, the \textit{availability} aspect (real-time performance) of the system, which is crucial for safety (e.g., causing vehicle collision~\cite{ndss:2022:ziwen:planfuzz}) has been relatively underexplored, especially for the complete camera-based AD perception pipeline. While some existing AD security analysis has studied availability in object detection~\cite{shapira2023phantom, chen2023overload}, they do not encompass the entire AD perception since usually, object detection is a part of the AD perception~\cite{jia2020fooling}. In addition, in the Cyber-Physical System area,
it is widely recognized that small component
level errors do not necessarily lead to system-level effects~\cite{shen2022sok, Wang_2023_ICCV}.
Thus, these studies leave a critical research gap: \textit{their proposed attack strategies may not be effective enough to conduct system-level effects in end-to-end AD systems}. As we demonstrate later, existing attacks targeting only object detection do not consistently produce highly potent system-level effects due to lack of entire AD perception consideration.

To fill in this critical research gap, in this paper, we are the first to study availability-based adversarial attacks across the entire camera-based AD perception pipeline, which includes both object detection and object tracking. Our proposed novel attack framework, \system, is designed to increase the latency of camera-based AD perception. Instead of solely targeting object detection, which might not yield potent system-level effects due to the limited increase of the latency, we realized the untapped potential of object tracking response time to generate a much more effective latency attack. To illustrate, an attacker focusing only on object detection might attempt to dramatically increase the number of proposed bounding boxes~\cite{chen2023overload}. Since object tracking stages might filter out a majority of these boxes and in common object detection post-processing steps~\cite{glenn_jocher_2020_4154370,zhang2021fairmot}, the maximum number of detection is provided to ensure their performance. Thus, the effectiveness of these attack is limited. Due to the importance of object tracking, 
we first perform availability attack surface analysis by
theoretically analyzing the time complexity of the state-of-the-art representative tracking algorithms, i.e., tracking-by-detection.
Then, we propose a two-stage attack strategy and formulate the attack as an optimization problem, shown in Fig.~\ref{fig_1}. Additionally, our novel loss function designs, encompassing score loss, bounding box area loss, and feature match loss, fully leverage the entire tracking-by-detection pipeline to generate effective latency-based attack.

Our experimental evaluation of \system targets four state-of-the-art camera-based AD perception pipelines. We find that \system, when compared with existing object detection latency attacks, provides significant improvements in latency under comparable perturbation levels. For instance, \system on average induces latency 2.9 times more than that for existing approaches~\cite{chen2023overload, shapira2023phantom}. We also demonstrate the system-level effects of our \system using Baidu Apollo~\cite{Apollo2017} LGSVL~\cite{rong2020lgsvl} AD simulator. The results show that \system induces a 70\% vehicle crash rate in two representative AD scenarios while existing methods achieve only a 30\% rate.  Demo videos are at the project website: \url{https://sites.google.com/view/cav-sec/slowtrack}

To sum up, our contributions are as follows:
% \vspace{-\topsep}
\begin{itemize}
\item We are the first to study availability-based adversarial attacks considering the entire AD
perception pipeline and find that previous object detection-based latency attack strategies may not induce potent system-level effects.
\item We propose a novel attack framework \system to systematically generate the latency adversarial attacks on camera-based AD perception by designing a two-stage attack strategy and proposing three novel loss functions.
\item \system is tested on four popular camera-based AD perception pipelines across different hardware, showing increase in latency and boost in system-level effects.

\end{itemize}

\section{Background and Related Work}
\label{sec:liturature}

\hspace*{10pt}{\bf Camera-based AD perception.} In the AD system, 
camera-based perception is primarily constituted by object detection and multi-object tracking (MOT)~\cite{Apollo2017, kato2018autoware}. This process aims to identify objects in each image frame and track their movement over time~\cite{jia2020fooling}.
\textit{Tracking-by-detection} has become
the dominant MOT paradigm~\cite{zhang2022bytetrack} and is widely used in industry-grade full-stack AD systems such as Baidu Apollo~\cite{Apollo2017} and Autoware.AI~\cite{kato2018autoware}.
It incorporates a detection module, a data association module, and a tracker management module. 
The detection module identifies objects in an image, noting their location, confidence, class scores, as well as other features for later data association. Data association then compares these detection with existing trackers based on features such as location and appearance, matching them based on similarity.

% While different 
Tracking-by-detection, despite varying in matching strategies, shares a similar tracking management~\cite{jia2020fooling} to build and delete the moving trajectories, called trackers, and mark trackers and detection boxes as different states.
Specifically, unmatched detection boxes are marked as unconfirmed and will be deleted unless they are continuously detected for \textit{H} frames. Matched trackers are marked as re-find or remain activated depending on the trackers' previous states, while unmatched trackers are marked as lost, and will be deleted if no objects are associated with them for \textit{R} frames. All of these trackers involved in matching constitute tracker$\_$pool and trackers with activated states are outputs.

Prior works~\cite{jia2020fooling, ma2023wip, shen2022sok} show that 
MOT poses a general challenge to cause AD system-level attack impact for existing attacks that target object detection since MOT is designed to be robust against errors in object detection. 
Given this challenge, our work introduces an innovative latency attack against the most representative and popular MOT (tracking-by-detection) and demonstrates the heightened system-level attack effects in AD.

{\bf Adversarial attack in AD.}
DNNs are vulnerable to
adversarial attacks~\cite{carlini2017towards, sato2023intriguing, luo2022infrastructure}, which are maliciously crafted samples to force DNNs to misbehave. Various prior works have explored the adversarial attacks in AD~\cite{cao20203d, sato2020hold, sato2021wip, sato2020security, muller2022physical, dipalma2021security}. 
While a majority of these attacks target integrity, our research concentrates on availability, which is another critical problem in AD~\cite{ndss:2022:ziwen:planfuzz}.
Although some attack works study availability~\cite{liu2023slowlidar, chen2023overload, shapira2023phantom, wang2021daedalus}, none of them consider the whole AD perception pipelines, which leads to suboptimal system-level effects in AD~\cite{jia2020fooling}.

{\bf Availability-based latency attack.} Availability-based latency attack can induce delays in the outputting function.
Such adversarial methods, when applied to DNN, have been investigated recently~\cite{shapira2023phantom, chen2023overload, liu2023slowlidar}.
However, their oversight of MOT within AD perception restricts their potential to achieve potent system-level effects.
Thus, in this work, we perform the first availability-based latency attack on the whole AD perception to significantly boost system-level effects.

\section{Methodology}
\label{sec:method}
\subsection{Availability Attack Surface Analysis}

\hspace*{10pt} To understand the vulnerability of the tracking-by-detection paradigm to latency attacks, we analyze the time complexity of the main three key steps (detection, data association, and tracker management) presented in Algorithm \ref{algo:byte}. The time consumption of the image preprocessing and backbone network of the detector hinges upon the computational dimension~\cite{shumailov2021sponge} and the number of computations \cite{hong2020panda,haque2020ilfo}.
Given that the dimensions of the input images are unchanged and the majority of the detection model activation values are inherently non-zero for the most of images \cite{shapira2023phantom}, we do not prioritize the time complexity of this segment.

The boxes obtained by the detection module usually need to be filtered before being passed to subsequent modules. Prevailing filtration techniques encompass non-maximum suppression (NMS) and score filtering with time complexity of $O(n^{2})$ and $O(n)$ respectively, where $n$ denotes the number of outputs from the detection network. However, since most tracking algorithms~\cite{zhang2021fairmot} confine the maximum number of detection to $\mathcal{|D|}_{max}$ , the maximum time complexities of these are $O(\mathcal{|D|}_{max}^{2})$ and $O(\mathcal{|D|}_{max})$.
Then, data association matches the reserved detection boxes and tracker$\_$pool with features, and the time complexity of this process is $O(mn^{\prime})$, where $m$ represents the number of trackers in tracker$\_$pool and $n^{\prime}$ represents the number of reserved detection boxes. In the tracker management module, trackers are created and deleted according to the matching results and are marked with different states with a time complexity of $O(m)$. Meanwhile, the velocity and location of the trackers are also updated with a time complexity of $O(m)$.

\begin{algorithm}[h!]
\SetAlgoLined
\DontPrintSemicolon
\SetNoFillComment
\footnotesize
\KwIn{A video sequence $\texttt{V}$; object detector $\texttt{Det}$; detection filter {$F$}; reserved age {$R$}}
\KwOut{Activated trackers $\mathcal{T}$ of the video}

Initialization: $\mathcal{T} \leftarrow \emptyset$\;
\For{frame $f_k$ in $\texttt{V}$}{
	\tcc{detection module}
	$\mathcal{D}_k \leftarrow \texttt{Det}(f_k)$ \;
        \textcolor{codegreen}{$\downarrow$ $O(n^{2})$  +  $O(n)$} \;
	$\mathcal{D}_{reserved} \leftarrow {F}({D}_k)$ \;
        \BlankLine	
        \textcolor{codegreen}{$\downarrow$ $O(m)$} \;
	\tcc{predict locations of trackers}
	\For{$t$ in $\mathcal{T}$}{
	$t \leftarrow \texttt{KalmanFilter}(t)$ \;
	}
	
    \BlankLine
    \BlankLine
    \textcolor{codegreen}{$\downarrow$ $O(mn^{\prime})$} \;
	Associate $\mathcal{T}$ and $\mathcal{D}_{reserved}$ using \texttt{Similarity}\;
     \textcolor{codegreen}{$\downarrow$ $O(m+n^{\prime})$} \;
	$\mathcal{D}_{matched} \leftarrow \text{matched object boxes from } \mathcal{D}_{reserved}$ \;
	$\mathcal{T}_{matched} \leftarrow \text{matched tracks from } \mathcal{T}$ \;
     $\mathcal{T}_{unmatched} \leftarrow \text{unmatched tracks from } \mathcal{T}$ \;
	
     \BlankLine
	\BlankLine
	\tcc{process unconfirmed tracks}
    \For{$t$ in $\mathcal{T}_{unconfirmed}$}{
	\If{$t {\in} \mathcal{T}_{matched}$}{
	$\mathcal{T} \leftarrow  \mathcal{T} \cup \{t\}$ \;
	}
 	\Else{
	{$\mathcal{T}_{unconfirmed} \leftarrow  \mathcal{T}_{unconfirmed} \setminus \{t\}$ \;}
	}
	}
    \BlankLine
    \tcc{add lost tracks}
     \For{$t$ in $\mathcal{T}_{unmatched} $}{
        $\mathcal{T}_{lost} \leftarrow  \mathcal{T}_{lost} \cup \{t\}$ \;}
	\BlankLine
        \tcc{reactivate tracks}
    \For{$t$ in $\mathcal{T}_{lost}\cap\mathcal{T}_{matched}$}{
	$\mathcal{T}_{lost} \leftarrow  \mathcal{T}_{lost} \setminus \{t\}$ \;
        $\mathcal{T} \leftarrow  \mathcal{T} \cup \{t\}$ \;
	}
	\tcc{delete unmatched tracks}
 \For{$t$ in $\mathcal{T}_{lost}$}{
	\If{$t_{lost\_time} > {R}$}{
	$\mathcal{T}_{lost} \leftarrow  \mathcal{T}_{lost} \setminus \{t\}$ \;
	}
 	\Else{
	{$t_{lost\_time}=t_{lost\_time}+{1}$} \;
	}}
 	\tcc{initialize new tracks}
    \For{$d$ in $\mathcal{D}_{matched}$}{
	$\mathcal{T}_{unconfirmed} \leftarrow  \mathcal{T}_{unconfirmed} \cup \{d\}$ \;
	}

 }
Return: $\mathcal{T}$
\caption{Tracking-by-detection}
\label{algo:byte}
\end{algorithm}

\subsection{Threat Model, Formulation, and Attack Overview}

\hspace*{10pt} {\bf Threat model.} Our attack method assumes white-box settings for the detection model, wherein both its architecture and parameters are known. For the tracking, we do not force the attackers to know the specific parameters and implementation details, which is a similar threat model as prior latency attack works~\cite{shapira2023phantom,chen2023overload}.

{\bf Formulation.} Detectors often restrict the maximum number of detection boxes. This constraint results in a worst-case time complexity for the detection module of $O(\mathcal{|D|}_{max}^{2})$ , effectively reducing the impact of previous latency attacks on object tracking~\cite{chen2023overload, shapira2023phantom}.
Thus, we propose an attack methodology that focuses on increasing the latency of the subsequent tracking stage under the constraint of limiting the maximum number of detection boxes. For a camera-based perception pipeline $P$, given the original image $x$, the attack goal is to craft an adversarial example $x^*$ to maximize the tracking pipeline latency, while keeping the added adversarial perturbations imperceptible. We formulate it as the following optimization problem:
\begin{equation}
	\label{eq:optatt}
\mathop{\arg\max}\limits_{x^*}T(P(x^*))
\quad
	\mathrm{s.t.} \,\, \left\{
	\begin{array}{l}
		{\mathcal{|D|}_{max}}=N \\
		\Delta (x^*, x) \leq \epsilon
	\end{array}
	\right.
\end{equation}
, where $T$ indicates the time function. Since the worst-case time complexity of the detection stage is $O(\mathcal{|D|}_{max}^{2})$, we can only increase the latency by increasing the number of trackers $m$. 
Our attack strategy is designed to create detection boxes that exploit vulnerabilities in the tracker management module of AD.
Specifically, for avoiding the temporary loss of objects in consecutive video frames due to occlusion, etc., the lost tracker will not be deleted immediately until the object is lost for $R$ consecutive frames. Leveraging this mechanism, we are able to inject more and more trackers into the tracking module by strategically creating detection boxes through carefully designed perturbations.
The detection boxes that appear in each frame are not associated with existing trackers and cause new tracking boxes to be created. As a result, the worst-case time complexity of the tracking module under our attack method is $O(R\mathcal{|D|}_{max}^{2})$, where $m = R\mathcal{|D|}_{max}$ and $n' = |D|_{max}$. Thus, it can lead to more computation cost than detection attacks in prior works..

\begin{figure}[!t]
\centering
\includegraphics[width=0.92\linewidth]{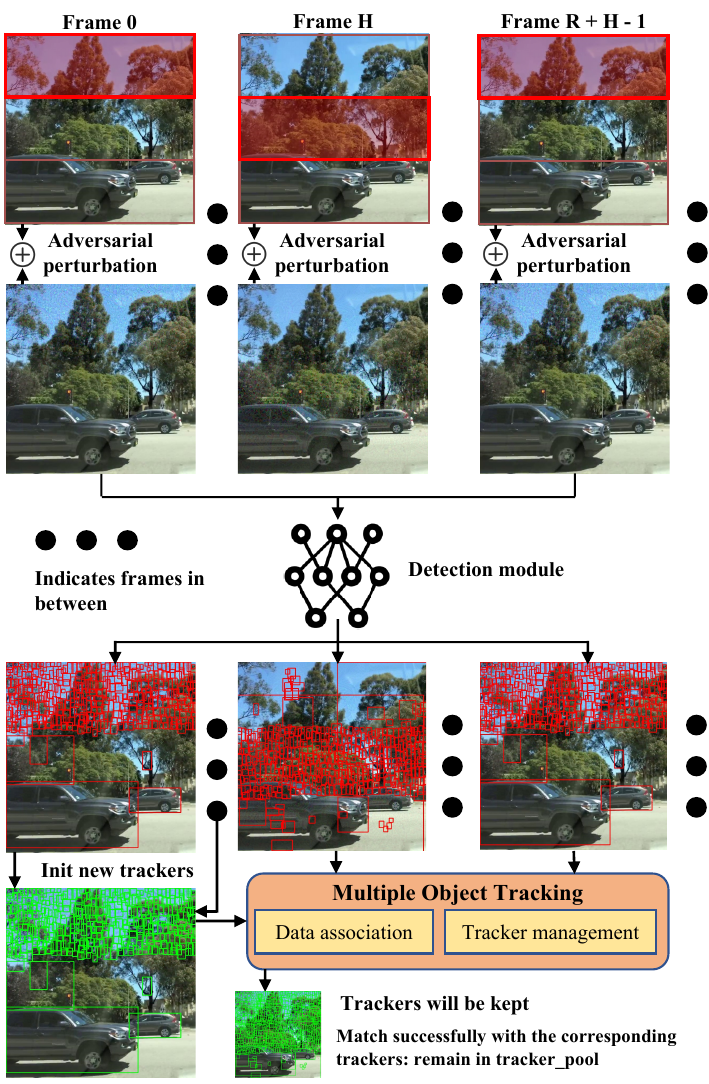}
\caption{Overview of our \system attack.}
\label{fig_1}
% \vspace{-0.6cm}
\end{figure}

{\bf Overview.}
In this paper, we propose \system, the first adversarial attack maximizing the latency of the whole camera-based perception pipeline, leveraging object tracking, which can significantly increase the latency of AD perception under the constraint shown in Eq~(\ref{eq:optatt}). The two attack stages of \system are: 1) in the attack initialization stage, make detection boxes created as many new trackers as possible, which requires detection boxes not to be associated with existing trackers, 2) make the lost trackers re-found before they are deleted, and kept in track$\_$pool, which requires detection boxes to be associated with corresponding trackers. Thus, we need to construct sets of detection boxes with the same matching features and make these sets of boxes appear or disappear in the corresponding video frames using adversarial attacks. To our best knowledge, representative tracking algorithms always use motion-based features for association. Thus, we divide the images into different regions and use the candidates in these regions as sets, which facilitates the inter-association of detection boxes within sets and the disassociation of detection boxes between sets. 

The overview of \system is in Fig.~\ref{fig_1}. Assuming that we divide the image into 3 regions, in frame 0, we select the top region and make the detection boxes appear in it, which are initialized as new track boxes. Similarly, in frame $H=1$ we make the fake detection boxes appear in the middle region, not associated with the existing track boxes, and make the tracker management module create new track boxes, while the track boxes in frame 0 do not disappear. These frames are not associated with existing frames, making the tracker management module create new trackers, and the trackers initialized in frame 0 are not deleted, allowing to inject $\mathcal{|D|}_{max}$ trackers into the track$\_$pool. Meanwhile, we make the detection boxes of the corresponding region reappear before the $R$ frames to keep the trackers not deleted. Since the tracker management module only initializes the detection boxes that detected $H$ frames consecutively, our attack strategy can be formulated as a scheduling problem: divide the image into $K$ regions, each of which needs to be selected for $H$ frames consecutively in the initialization stage, and then be selected once more before $R$ frames. We use the greedy algorithm with the maximum value of $K$: $R-H+1$ (when $R-K=H-1$, no region can be added).

{\bf Attack Strategy Generating.} Algorithm \ref{algo:byte} outlines the full process of Tracking-by-detection.
We analyze the representative tracker management module, which is also used in some joint-tracking algorithms and propose the attack strategy generation algorithm shown in Algorithm \ref{algo:stra}. Specifically, we use a greedy algorithm to generate attack strategy to continuously inject different regions of detection boxes into the tracker management module. Such an attack strategy can also be generalized to different joint-tracking algorithm.

\begin{algorithm}[t]
\SetAlgoLined
\DontPrintSemicolon
\SetNoFillComment
\footnotesize
\KwIn{A video sequence $\texttt{V} = [v_0, v_1, ..., v_{K-1}]$; reserved age {$R$}; hit count {$H$}}
\KwOut{attack strategy $\mathcal{S}$}
\tcc{$region\_idx$ is selected image region}
Initialization: $\mathcal{S} \leftarrow \emptyset;\text{  }region\_idx = 0;\text{  } n = 0$\;

% $$\;
% \;
% $Re \leftarrow \emptyset$\;
\While{$n<K$}{
\tcc{$Re$ is the next time each region needs to be reactivated}
	\If{$n == 0$}{
	$\mathcal{S} \leftarrow  \mathcal{S} \cup \{region\_idx\}$ \;
        $Re_{region\_idx} = n+R+1$ \;
        {${region\_idx} = {region\_idx}+{1}$} \;
        % {${n} = {n}+{1}$} \;
	}
 	\Else{
            $Re_{min},idx \leftarrow \text{minimum value and index in } {Re}$ \;
    	\If{$Re_{min}-n \textless {H}$}{
    	$\mathcal{S} \leftarrow  \mathcal{S} \cup \{idx\}$ \;
            $Re_{idx} = n+R+1$ \;
            % {${n} = {n}+{1}$} \;
    	}
     	\Else{
              \For{$i=1$ to $H$}{
        	$\mathcal{S} \leftarrow  \mathcal{S} \cup \{region\_idx\}$\;
        	}
            $Re_{region\_idx} = n+R+1$ \;
            {${region\_idx} = {region\_idx}+{1}$} \;
            {${n} = {n}+{H-1}$} \;
            }
    	}
        {${n} = {n}+{1}$} \;
	}
	
Return: $\mathcal{S}$
\caption{Generate Attack Strategy}
\label{algo:stra}
\end{algorithm}

\subsection{Loss Function Design}
\hspace*{10pt}We compose novel loss functions to achieve the attack above, which consists of three components: (a) score loss, (b) bounding box area loss, and (c) feature match loss.

{\bf Score loss.} To boost the number of selected boxes, it is necessary to raise the number of prediction candidates that bypass the detection filter, which selects candidates based on their confidence scores.
Thus, to increase confidence score of selected candidates $\mathcal{C}_{sel}$, we propose a novel score loss:
\begin{center}
${{\mathcal{L}_{score} = {\frac{1}{|\mathcal{C}_{sel}|}} \sum\limits_{{c}\in\mathcal{C}_{sel}}\mathop{\max}((T_{conf}-c_{conf}),\lambda)}}$
\end{center}
where $T_{conf}$ represents the filtering confidence threshold set by the detection model, $c_{conf}$ represents the confidence scores of the object detector, and $\lambda$ is a hyper-parameter.

{\bf{Bounding box area loss.}} To make more candidates to be reserved in the NMS employed by some detection filters, we need to compress the dimensions of the boxes to reduce the IOU between the candidates. This is expressed as:
\begin{center}
${{\mathcal{L}_{area} = {\frac{1}{|\mathcal{C}_{sel}|}} \sum\limits_{{c}\in\mathcal{C}_{sel}}(\frac{b^w_c \cdot b^h_c}{S_W \times S_H})^2}}$
\end{center}
where the bounding box is ${b}$, with ${b^w}$ and ${b^h}$ being its width and height. ${S_W}$ and ${S_H}$ are the width and height of the input image. This loss is added only when the filter contains NMS.

{\bf{Feature matching loss.}}
To successfully match the selected detection box with corresponding lost tracker so that the lost tracker is re-found before being deleted, the feature distance for the data association module needs minimizing.
\begin{center}
${\mathcal{L}_{match} = \Psi(\mathcal{T}_i, F^{\prime}(\mathcal{C}_{sel}))}$
\end{center}
where $\Psi$ is the feature distance function in data association, $\mathcal{T}_i$ represents i-th set of trackers, and $F^{\prime}$ represents detection filters without score threshold filtering.

Pairwise computation for obtaining feature distances could be expensive if feature extraction is complex or if the number of detection boxes is too large. Therefore, we propose a less computationally intensive method to make the corresponding trackers and detection frames match successfully. 
The images that need to appear with the same batch of detection boxes use the same perturbation. However, this makes the perturbation accumulation more obvious and the result of the attack decreases. Therefore, for balance, we use the universal perturbation method in the attack initialization stage and feature matching loss after that.
Finally, the adversarial loss is represented by:
\begin{equation}
\label{eq:total-loss}
    {\mathcal{L}_{adv} = \lambda}_1\mathcal{L}_{score}+{\lambda}_2\mathcal{L}_{area}+{\lambda}_3\mathcal{L}_{match}
\end{equation}
Similar to existing works~\cite{carlini2017towards}, to make the perturbation invisible, we constrain the $\mathcal{L}_2$ norm: 
\begin{center}
$\mathop{\min}\limits_{x^*}\mathcal{L}_{adv}+{\mu}{\Vert x^*-x \Vert_2}$
\end{center}
where $\mu$ is the hyper-parameter and $x$ is the original image.

\section{Experiments}
\label{sec:eval}

\begin{figure}[!t]
\centering
\includegraphics[width=0.96\linewidth]{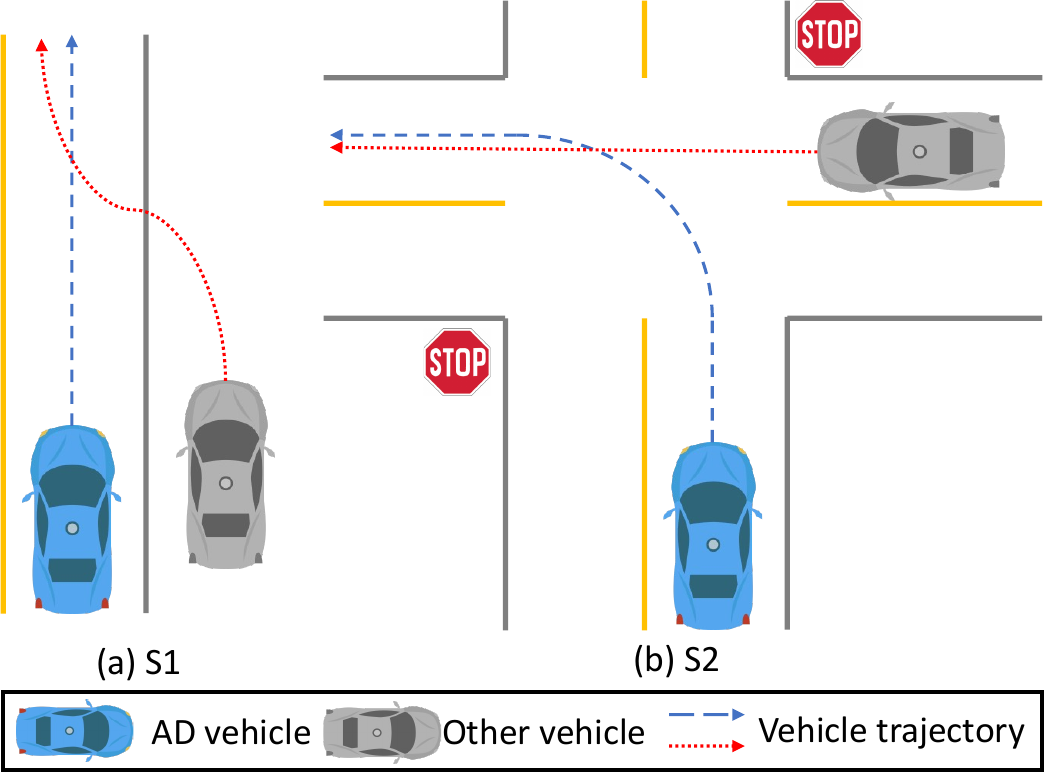}

\caption{Two scenarios (S1 and S2) for our simulation evaluation setup on the system-level effects.}
\label{fig:scenario}
% \vspace{-0.4cm}
\end{figure}

\begin{table*}[t]
\tabcolsep 0.045in
\small

    \centering
    \begin{tabular}{ccccccccccccccc}

    \toprule
    & & & \multicolumn{4}{c}{PS~\cite{shapira2023phantom}} & \multicolumn{4}{c}{Overload~\cite{chen2023overload}} & \multicolumn{4}{c}{\system}\\
    \cmidrule(lr){4-7}
    \cmidrule(lr){8-11}
    \cmidrule(lr){12-15}
    
    \multirow{-2}{*}{Model} & \multirow{-2}{*}{Dataset} &  \multirow{-2}{*}{Hardware} & R-Track & R-Lat & \#Track & $ \mathcal{L}_2$ & R-Track & R-Lat & \#Track & $\mathcal{L}_2$ & R-Track & R-Lat & \#Track & $\mathcal{L}_2$\\
    \midrule

    & & Titan V & 73.1 & 6.2 &  &  & 156.1 & 12.1 & & & \textbf{247.0} & \textbf{17.5} \\ 
    & & 2080 Ti & 64.9 & 11.7 &  &  & 205.1 & 34.7 & & & \textbf{310.3} & \textbf{49.4} \\
    & \multirow{-3}{*}{BDD} & 3090 & 76.4 & 6.2 &  \multirow{-3}{*}{69.1}  &  \multirow{-3}{*}{0.042} & 186.3 & 13.8 & \multirow{-3}{*}{94.6}  &  \multirow{-3}{*}{0.011}& \textbf{338.3} & \textbf{23.0} &  \multirow{-3}{*}{\textbf{141.8}}  &  \multirow{-3}{*}{\textbf{0.010}} \\
    % & \multirow{-4}{*}{BDD} & V100 \\
    \cmidrule(lr){2-15}
    & & Titan V & 37.5 & 3.8 &  &  & 93.0 & 8.7 & & & \textbf{208.4} & \textbf{18.5} \\
    & & 2080 Ti & 33.2 & 8.6 & & & 82.6 & 19.3 & & & \textbf{225.5} & \textbf{50.9} \\
    \multirow{-6}{*}{SORT (Y5)} & \multirow{-3}{*}{MOT17} & 3090  & 39.8 & 3.7 &  \multirow{-3}{*}{32.4}  &  \multirow{-3}{*}{0.041} & 111.1 & 9.8 & \multirow{-3}{*}{47.0}  &  \multirow{-3}{*}{\textbf{0.013}}& \textbf{283.9} & \textbf{25.7} &  \multirow{-3}{*}{\textbf{73.5}}  &  \multirow{-3}{*}{\textbf{0.013}} \\
    % \multirow{-8}{*}{YOLOv5 + SORT}& \multirow{-4}{*}{MOT} & V100 \\

    \midrule

    & & Titan V & 517.7 & 10.4 & & & 1505.9 & 29.1 & & & \textbf{2647.6} & \textbf{51.3} \\
    & & 2080 Ti & 390.1 & 8.7 & & & 1258.7 & 26.7 & & & \textbf{2260.7} & \textbf{48.1} & \\
    & \multirow{-3}{*}{BDD} & 3090 & 236.0 & 3.8 & \multirow{-3}{*}{401.1} & \multirow{-3}{*}{0.036} & 836.4 & 13.3 & \multirow{-3}{*}{939.9} & \multirow{-3}{*}{0.029} & \textbf{1572.9} & \textbf{25.2} & \multirow{-3}{*}{\textbf{1334.9}} & \multirow{-3}{*}{\textbf{0.028}} \\
    % & \multirow{-4}{*}{BDD} & V100 \\
    \cmidrule(lr){2-15}
    & & Titan V & 67.4 & 8.6 & & & 181.7 & 21.5 & & & \textbf{341.0} & \textbf{41.5} \\
    & & 2080 Ti & 49.3 & 7.1 & & & 149.0 & 19.9 & & & \textbf{279.5} & \textbf{38.4} \\
    \multirow{-6}{*}{FairMOT} &\multirow{-3}{*}{MOT17}  & 3090 & 32.4 & 3.2 & \multirow{-3}{*}{48.7} & \multirow{-3}{*}{0.036} & 108.5 & 10.1 & \multirow{-3}{*}{106.2} & \multirow{-3}{*}{\textbf{0.026}} & \textbf{220.7} & \textbf{21.3} & \multirow{-3}{*}{\textbf{159.0}} & \multirow{-3}{*}{\textbf{0.026}}\\
    % \multirow{-8}{*}{FairMOT}& \multirow{-4}{*}{MOT} & V100 \\

    \midrule

    & & Titan V & 40.2 & 2.5 & & & 173.0 & 9.2 & & & \textbf{290.0} & \textbf{14.7}\\
    & & 2080 Ti & 39.5 & 2.3 & & & 184.0 & 9.0& & & \textbf{266.4} & \textbf{12.5}\\
    & \multirow{-3}{*}{BDD} & 3090 & 57.8 & 3.0 & \multirow{-3}{*}{79.2} & \multirow{-3}{*}{0.032} & 217.7 & 10.1 & \multirow{-3}{*}{224.7} & \multirow{-3}{*}{0.027} & \textbf{341.4} & \textbf{15.1} & \multirow{-3}{*}{\textbf{307.0}} & \multirow{-3}{*}{\textbf{0.022}} \\
    % & \multirow{-4}{*}{BDD} & V100 \\
    \cmidrule(lr){2-15}
    & & Titan V & 29.0 & 1.9 & & & 95.0 & 5.4 & & & \textbf{173.0} & \textbf{9.8}\\
    & & 2080 Ti & 26.9 & 1.8 & & & 97.7 & 5.4 & & & \textbf{168.7} & \textbf{9.5}\\
    \multirow{-6}{*}{ByteTrack}&\multirow{-3}{*}{MOT17} & 3090 &  45.2 & 2.4 & \multirow{-3}{*}{38.0} & \multirow{-3}{*}{0.030} & 115.2 & 5.9 & \multirow{-3}{*}{78.3} & \multirow{-3}{*}{0.025} & \textbf{204.0} & \textbf{10.5} & \multirow{-3}{*}{\textbf{130.1}} & \multirow{-3}{*}{\textbf{0.022}}\\
    % \multirow{-8}{*}{ByteTrack}& \multirow{-4}{*}{MOT} & V100 \\

    \midrule

    & & Titan V & 53.2 & 19.4 & & & 79.0 & 28.9 & & & \textbf{92.1} & \textbf{33.6}\\
    & & 2080 Ti & 57.6 & 19.6 & & & 89.1 & 30.6 & & & \textbf{103.6} & \textbf{35.5}\\
    &\multirow{-3}{*}{BDD} & 3090 & 61.0 & 22.1 & \multirow{-3}{*}{70.0} & \multirow{-3}{*}{0.033} & 93.9 & 34.1 & \multirow{-3}{*}{221.9} & \multirow{-3}{*}{0.029} & \textbf{135.3} & \textbf{49.7} & \multirow{-3}{*}{\textbf{280.8}} & \multirow{-3}{*}{\textbf{0.025}}\\
    % & \multirow{-4}{*}{BDD} & V100 \\
    \cmidrule(lr){2-15}
    & & Titan V & 31.7 & 14.0 & & & 42.1 & 18.5 & & & \textbf{52.2} & \textbf{23.0} \\
    & & 2080 Ti & 34.2 & 14.6 & & & 45.6 & 19.5 & & & \textbf{59.1} & \textbf{25.3}\\
    \multirow{-6}{*}{BoT-SORT}& \multirow{-3}{*}{MOT17} & 3090 & 44.0 & 19.9 & \multirow{-3}{*}{40.7} & \multirow{-3}{*}{0.034} & 47.2 & 21.1 & \multirow{-3}{*}{83.9} & \multirow{-3}{*}{\textbf{0.025}} & \textbf{69.4} & \textbf{30.8} & \multirow{-3}{*}{\textbf{127.6}} & \multirow{-3}{*}{\textbf{0.025}}\\
    % \multirow{-8}{*}{TransTrack}& \multirow{-4}{*}{MOT} & V100 \\

    \bottomrule
         
    \end{tabular}
    % \vspace{-0.2cm}
    \caption{Effectiveness results of tracking-stage and whole perception latency with number of trackers and average $\mathcal{L}_2$ norm in different models and hardware. \textbf{Bold} denotes the best results (i.e., highest R-Track, R-Lat, \#Track, and lowest $\mathcal{L}_2$) in each row.}
    \label{tab:main_res}
\end{table*}

\begin{table}[t!]
    \small
    \centering
    \begin{tabular}{clccc}
        \toprule 
        Model & Attack & Titan V & 2080 Ti & 3090 \\
        \midrule
        % Model & \multicolumn{3}{c}{SORT(Y5)} \\
        &PS & 211 & 199 & 160 \\
        &Overload & 406 & 411 & 337 \\
        \multirow{-3}{*}{SORT (Y5)}&\system & \textbf{847} & \textbf{1082} & \textbf{1018} \\
        \midrule
        % Model & \multicolumn{3}{c}{FairMOT} \\
        &PS & 417 & 359 & 259 \\
        &Overload & 970 & 914 & 685 \\
        \multirow{-3}{*}{FairMOT}&\system & \textbf{1848} & \textbf{1731} & \textbf{1726} \\
        \midrule
        % Model & \multicolumn{3}{c}{ByteTrack} \\
        &PS & 174 & 188 & 166 \\
        &Overload & 379 & 427 & 330 \\
        \multirow{-3}{*}{ByteTrack}&\system & \textbf{584} & \textbf{621} & \textbf{555} \\
        \midrule
        % Model & \multicolumn{3}{c}{BoT-SORT} \\
        &PS & 2395 & 2460 & 2372 \\
        &Overload & 3093 & 3195 & 2485 \\
        \multirow{-3}{*}{BoT-SORT}&\system & \textbf{3768} & \textbf{4101} & \textbf{3245} \\
        \bottomrule
    \end{tabular}
    \caption{Latency Time (ms) on MOT 17 dataset.}
    \label{tab:time}
\end{table}

% \begin{table}[t]
% \tabcolsep 0.03in
% \small
% % \renewcommand{\arraystretch}{0.75}
%     \caption{\system transferability evaluation on \#Track across different models with MOT17 dataset.}
%     \centering
%     \begin{tabular}{ccccc}

%     \toprule
%     \diagbox{Source}{Target} & SORT (Y5) & FairMOT & ByteTrack & BoT-SORT\\
%     \midrule
%     SORT (Y5) & 65.8 & -0.1 & 0.0 & -0.1\\
%     FairMOT & 0 & 159.0 & 0 & 0\\
%     ByteTrack & 0 & 0 & 130.1 & 0\\
%     BoT-SORT & 0 & 0 & 0 & 127.6\\

%     \bottomrule
         
%     \end{tabular}

%     \label{tab:tranfer}
% \end{table}

\begin{table}[t]
\tabcolsep 0.05in
\small

    \centering
    \begin{tabular}{ccccc}

    \toprule
     & w/o $\mathcal{L}_{score}$  & w/o $\mathcal{L}_{area}$ & w/o $\mathcal{L}_{match}$ & \system\\
    \midrule
    R-Track & 0.0 & 153.2 & 171.4 & 225.5\\
    R-Lat & 0.0 & 34.0 & 38.3 & 50.9\\
    \#Track & 0.1 & 63.9 & 66.9 & 73.5 \\

    \bottomrule
         
    \end{tabular}
    % \vspace{-0.4cm}
    \caption{Ablation study for loss designs ($\mathcal{L}_{score}$, $\mathcal{L}_{area}$, and $\mathcal{L}_{match}$ in Eq. (\ref{eq:total-loss})) on R-Track, R-Lat, and \#Track with SORT (Y5) and MOT17 using 2080 Ti. w/o: without}
    \label{tab:ablation}
\end{table}

\begin{table}[t!]
\tabcolsep 0.02in
\small

    \centering
    \begin{tabular}{ccccccc}

    \toprule
    & \multicolumn{3}{c}{S1} & \multicolumn{3}{c}{S2}\\
    \cmidrule(lr){2-4}
    \cmidrule(lr){5-7}
    \multirow{-2}{*}{Model} & PS & Overload & \system & PS & Overload & \system\\
    
    % & & & \multicolumn{4}{c}{PS~\cite{shapira2023phantom}} & \multicolumn{4}{c}{Overload~\cite{chen2023overload}} & \multicolumn{4}{c}{SlowTrack}\\
    % \cmidrule(lr){4-7}
    % \cmidrule(lr){8-11}
    % \cmidrule(lr){12-15}
    % \multirow{-2}{*}{Model} & \multirow{-2}{*}{Dataset} &  \multirow{-2}{*}{Hardware} & ROI-Pro & ROI-Lat & $\Delta \mathcal{L}_1$ & $\Delta \mathcal{L}_2$ & ROI-Pro & ROI-Lat & $\Delta \mathcal{L}_1$ & $\Delta \mathcal{L}_2$ & ROI-Pro & ROI-Lat & $\Delta \mathcal{L}_1$ & $\Delta \mathcal{L}_2$\\
    
    \midrule
    SORT(Y5) & 10\% & 30\% & \textbf{100\%} & 20\% & 40\% & \textbf{90\%} \\
    FairMOT & 20\% & 40\% & \textbf{100\%} & 20\% & 40\% & \textbf{100\%} \\
    ByteTrack & 0\% & 40\% & \textbf{80\%} & 0\% & 40\% & \textbf{90\%} \\
    BoT-SORT & 40\% & 50\% & \textbf{100\%} & 50\% & 50\% & \textbf{100\%} \\
    % & 262 & 699 & 1295& 359 & 914 & 1731& 188 & 427 & 621& 2460 & 3195 & 4101

    \bottomrule
         
    \end{tabular}
    % \vspace{-0.3cm}
    \caption{System-level evaluation (vehicle crash rate) with Baidu Apollo and LGSVL simulator. 10 runs for each cell.}
    \label{tab:simulation}
\end{table}

\subsection{Experimental Setup}

\hspace*{10pt} {\bf Datasets and models.}
We use the MOT17DET (MOT17)~\cite{milan2016mot16} and BDD~\cite{seita2018bdd100k} datasets to evaluate the proposed attacks. MOT17 includes 14 videos with more than 10K images and BDD contains 100K images with various attributes such as weather, scene, and time of day, resulting
in a diverse dataset. For MOT17, we use all the data while for BDD, we randomly select 10 videos.
%, which is the similar setup as existing work~\cite{jia2020fooling}. 
As for the models, we select the most representative perception models: SORT (Y5)~\cite{bewley2016simple} (a Kalman filtering-based MOT generally used in AD~\cite{Apollo2017, shen2022sok}, with YOLO v5~\cite{glenn_jocher_2020_4154370} as detector), FairMOT~\cite{zhang2021fairmot}, ByteTrack~\cite{zhang2022bytetrack}, and BoT-SORT~\cite{aharon2022bot}.
We use the suggested parameter for each model.
% \todo{add more discussion on models including tracking and ODs}

{\bf Evaluation metrics.} We design the metrics as follows:

% i.e., number of tracker increase and latency increase, we use the two metrics below:

\begin{center}
\small
$\textrm{R-Track} = \dfrac{\textrm{Track-Lat}(x^*) - \textrm{Track-Lat}(x)}{\textrm{Track-Lat}(x)}$

$\textrm{R-Lat} = \dfrac{\textrm{Total-Lat}(x^*) - \textrm{Total-Lat}(x)}{\textrm{Total-Lat}(x)}$

$\textrm{\#Track} = \dfrac{\textrm{Tracker\#}(x^*) - \textrm{Tracker\#}(x)}{\textrm{Tracker\#}(x)}$
\end{center}

\noindent where R-Track and R-Lat represent the rate of increase for the tracking latency and whole perception latency and \#Track represents the rate of increase for the number of tracker. To measure the imperceptibility, we use 
average $\mathcal{L}_2$ norm~\cite{carlini2017towards, zhang2020interpretable}. We define system-level effects metric as vehicle crash rate: $\dfrac{N_\textrm{crash}}{N_\textrm{total}}$, where $N_\textrm{crash}$ denotes the number of runs causing vehicle crashes and $N_\textrm{total}$ is the number of total runs.

{\bf Testing hardware.} Given that latency is intrinsically tied to the hardware device, we test \system on multiple hardware: TiTAN V, GeForce RTX 2080 Ti (shown to be used in real AD~\cite{shen2022sok}), and GeForce RTX 3090.

% Since the latency is highly correlated to the hardware device, we evaluate our attack with different hardware: TiTAN V, GeForce RTX 2080 Ti GPU, which is shown to be used in real AD vehicle~\cite{shen2022sok}, and GeForce RTX 3090 GPU.

{\bf Baselines comparison.} To our best knowledge, we are the first to propose a latency attack against whole camera-based AD perception pipeline, while the existing attacks focus on object detection alone. Thus, we select two representative latency attacks on object detection as our baselines: PS~\cite{shapira2023phantom} and Overload~\cite{chen2023overload}.

{\bf Simulation evaluation.} To study the system-level effects, we perform an end-to-end attack evaluation on Baidu Apollo~\cite{Apollo2017}, an industry-grade full-stack AD system, with LGSVL simulator~\cite{rong2020lgsvl}, a production-grade AD simulator. 
Our experiments are conducted on the Borregas Ave map and the Lincoln2017MKZ AD vehicle with default configuration.
To simulate our attack impact, we model the latency of the camera-based AD perception and inject it into the AD system. Due to the representativeness of SORT (Y5), we use its latency results tested on 2080 Ti GPU (used in genuine AD vehicles~\cite{shen2022sok}), as our latency modeling results.
Our evaluation focuses on two representative scenarios as shown in Fig.~\ref{fig:scenario}, where the blue vehicle is the victim AD vehicle and the blue and red lines are the trajectories of the two vehicles. S1 (Fig.~\ref{fig:scenario} (a)) is a common driving scenario for other vehicles to change the lane line and S2 (Fig.~\ref{fig:scenario} (b)) is another common driving scenario for the STOP sign-controlled intersection. We perform 10 runs on each scenario and compare \system with the two baselines.

\subsection{Experimental Results}

\hspace*{10pt}{\bf Effectiveness}. As shown in Table~\ref{tab:main_res}, we compare our \system with the baselines. \system can increase the number of tracker up to 1334.9 on FairMOT, which is much better than existing works: at most 939.9 on FairMOT. Especially, for the tracking stage, \system provides 453.8 times slowing on average compared to the existing works which only have 256.4 times for Overload and 89.1 times for PS. As for the latency of the whole camera-based AD perception, we find that \system can provide 28.4 times slowing down on average, which is 2.9 times more than the two existing works. Especially, for the practical SORT (Y5) on 2080 Ti, we observe 50.9 times slowing down, while existing works can only have 19.3 times at most. For the imperceptibility, our average $\mathcal{L}_2$ norm (around 0.021 on average) is very small compared to the baselines and existing adversarial attacks on integrity~\cite{zhang2020interpretable}. 

We measure the latency time in Table~\ref{tab:time}. 
For instance, while existing representative attack Overload~\cite{chen2023overload} can trigger 411 ms latency on SORT (Y5) and MOT17 dataset with 2080 Ti, \system can provide 1,082 ms, which is 1.6 times more than Overload. Thus, \system can significantly outperform the existing baselines on camera-based AD perception, with similar levels of imperceptibility.

{\bf Ablation study.}
The ablation study evaluating the designs of Eq.~(\ref{eq:total-loss}) is presented in Table\ref{tab:ablation}. In this study, each loss component is sequentially removed, and the attack is tested on the most practical SORT (Y5) tracking method and the MOT17 dataset, utilizing a 2080 Ti. The findings in Table~\ref{tab:ablation} highlight the indispensability of all three loss designs in achieving high attack effectiveness. Notably, the $\mathcal{L}_{score}$ performs as the most pivotal element in enhancing attack effects. In the absence of $\mathcal{L}_{area}$ and $\mathcal{L}_{match}$, the R-Lat experiences reductions of approximately 33\% and 25\%, respectively. Thereby, these results indicate the importance of integrating all three loss designs together.

\begin{figure}[!t]
\centering
\subfloat[Confidence]{\includegraphics[width=1.57in]{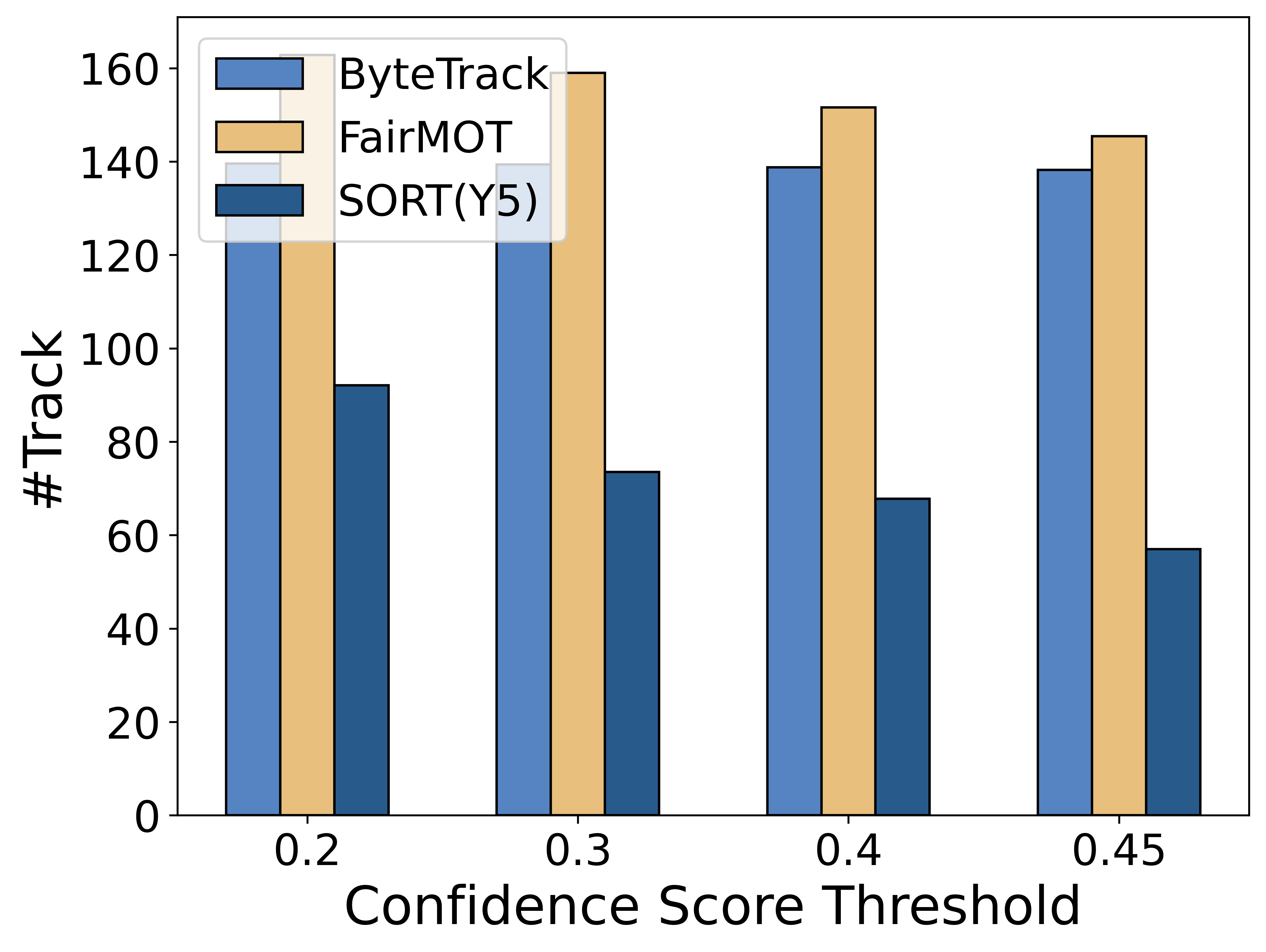}%
\label{fig_3a}}
% \vspace{-0.1cm}
\subfloat[Maximum Number]{\includegraphics[width=1.57in]{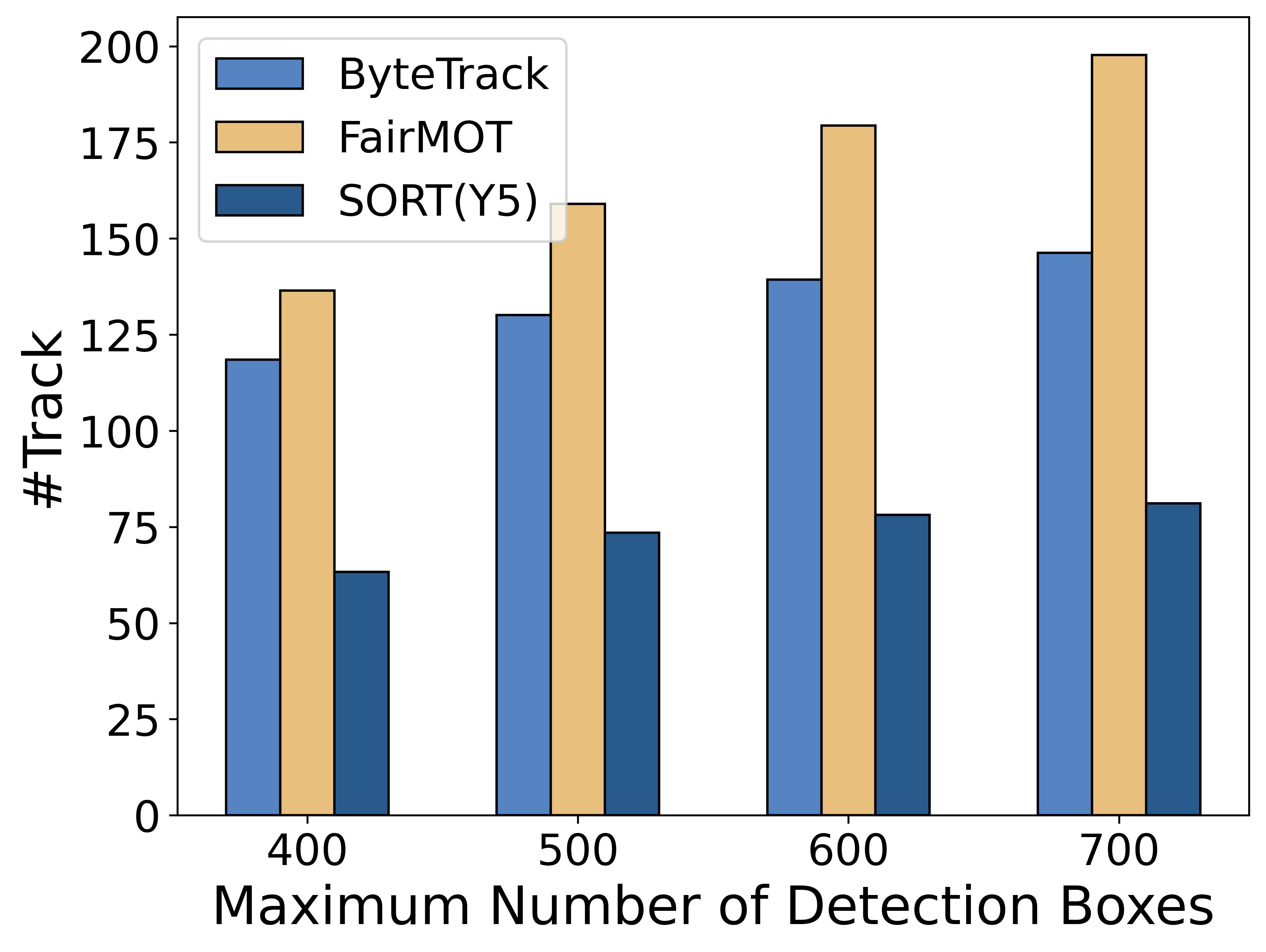}%
\label{fig_3b}}
\hfill
% \vspace{-0.3cm}
\subfloat[IOU Threshold for NMS]{\includegraphics[width=1.57in]{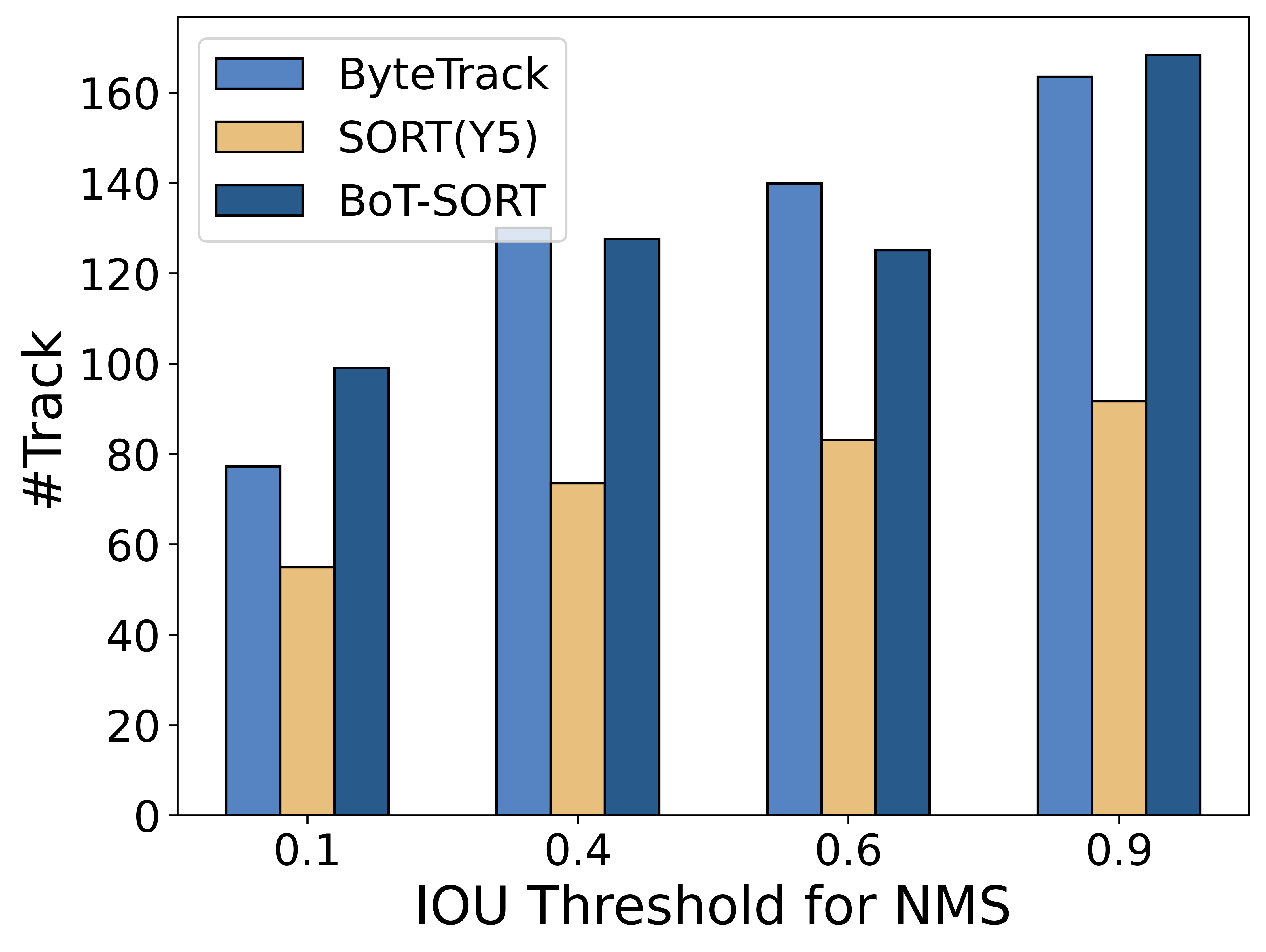}%
\label{fig_3c}}
\subfloat[IOU Threshold for Match]{\includegraphics[width=1.57in]{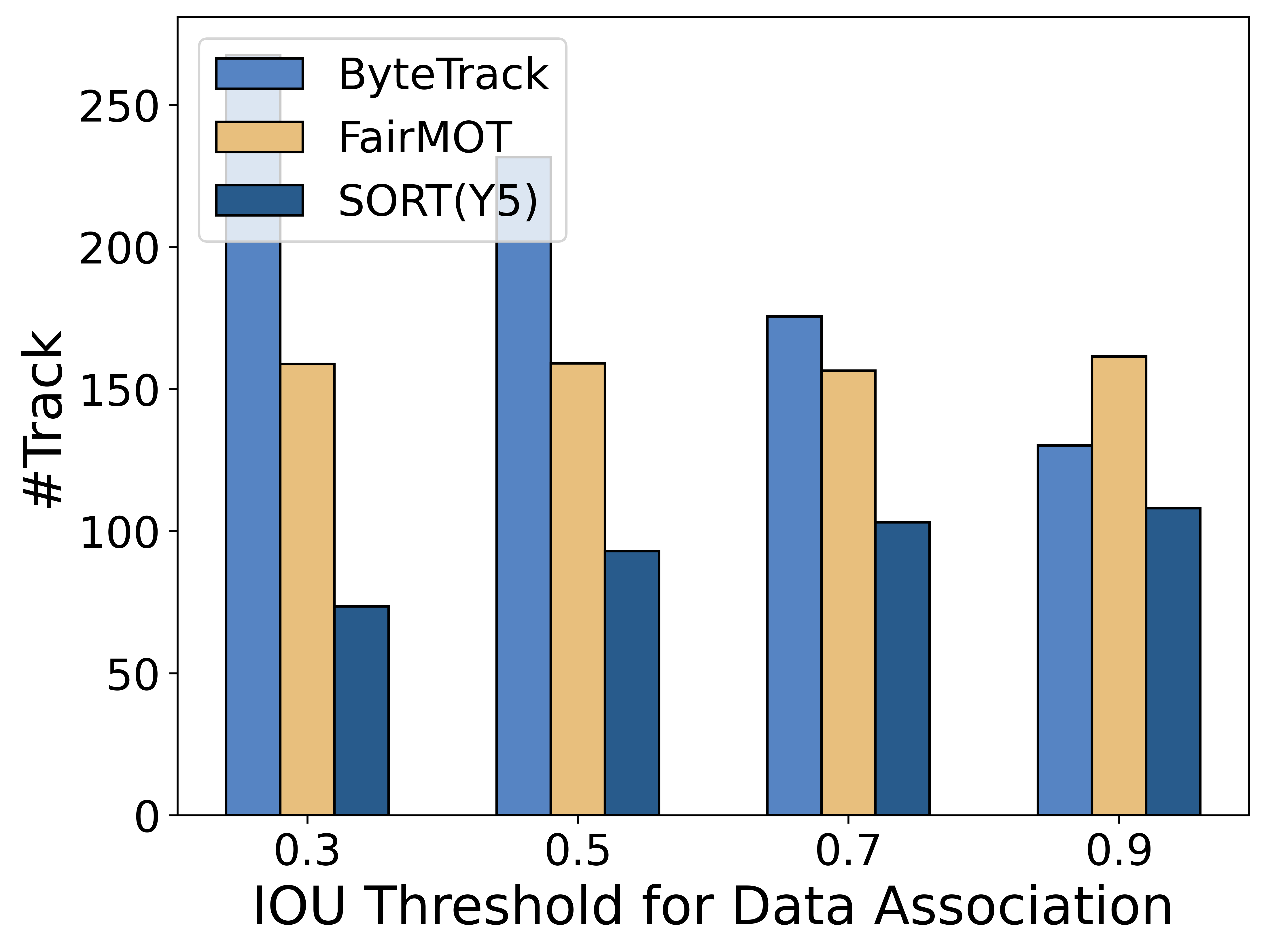}%
\label{fig_3d}}
% \vspace{-0.2cm}
\caption{Attack effectiveness under different thresholds.}
\label{fig:fig_3}
% \vspace{-0.3cm}
\end{figure}

{\bf Generality to different thresholds.} The effectiveness of \system is most impacted by different thresholds: confidence score threshold in object detection, maximum number of detection boxes, IOU threshold for NMS, and IOU threshold for data association. To evaluate their impact, we vary the thresholds with different values and measure the \#Track on the MOT17 dataset with 2080 Ti. The results are shown in Fig.~\ref{fig:fig_3}. \system can generally generate stable results across different threshold parameters, which indicates that setting better thresholds cannot fully defend against \system. For instance, setting the confidence score threshold to 0.2, the \#Track of SORT (Y5) is around 92.1 while setting the confidence score threshold to 0.4, the \#Track is around 67.8. Additionally, finetuning the thresholds usually will 
significantly reduce the benign performance. Thus, the results show the generality
of \system to different thresholds.

{\bf End-to-end simulation evaluation results.} As shown in Table~\ref{tab:simulation}, \system can achieve 95\% vehicle crash rate on average while for other attacks, they can only have around 30\%. Note that the vehicle crash rate in benign cases is always 0\%. The results demonstrate that the existing latency attacks on object detection alone cannot trigger sufficient latency to highly effectively cause vehicle crashes, which motivates our new attack design on entire perception. 
\system significantly improves the system-level effects.

\begin{table}[t]
\tabcolsep 0.04in
\small

    \centering
    \begin{tabular}{ccccccc}

    \toprule
    & \multicolumn{2}{c}{Titan V} & \multicolumn{2}{c}{2080 Ti} & \multicolumn{2}{c}{3090} \\
    
    \cmidrule(lr){2-3}
    \cmidrule(lr){4-5}
    \cmidrule(lr){6-7}
    \multirow{-2}{*}{Model} & R-Track & R-Lat & R-Track & R-Lat & R-Track & R-Lat\\
    
    \midrule
    SORT(Y5) & 0.032 & 0.001 & 0.106 & 0.007 & 0.459 & 0.007\\
    FairMOT & 0.402 & 0.004 & 0.963 & 0.010 & 0.835 & 0.005\\
    ByteTrack & 0.156 & 0.002 & 0.075 & 0.003 & 0.050 & 0.001\\
    BoT-SORT & 0.006 & 0.001 & 0.158 & 0.032 & 0.192 & 0.054\\

    \bottomrule
         
    \end{tabular}
    \caption{Variances of R-Track and R-Lat in repeated experiments (5 times) with different hardware on MOT17 dataset.}
    \label{tab:var}
    % \vspace{-0.6cm}
\end{table}

{\bf Running time variance.} Recognizing the potential variability of system latency due to many factors, we repeat our experiments 5 times~\cite{variance2017time} and measure the variance of R-Track and R-Lat. As shown in Table~\ref{tab:var}, the variance is negligible compared to the original values~\cite{liu2023slowlidar}.

\section{Discussion}
\label{sec:discussion}
\subsection{Physical-World Attack Realizability}
\hspace*{10pt} Our study investigates the runtime robustness of camera-based AD perception by adversarial attacks. Such attack is physically realizable as demonstrated in AttrackZone attack~\cite{muller2022physical}, which leverages the projector to project the noise-level adversarial attack in physical world at night. To improve our attack realizability, we take a small step forward to generate a patch-based adversarial attack which is generally demonstrated as a physical-world realizable attack~\cite{Wang_2023_ICCV, wang2022poster}.

{\bf Adversarial patch generation.} To formulate the patch into a patch $\delta$, we design the following method: 
\begin{center}
$\min\limits_{\delta} {E}_{x\sim\mathcal{X}}   \mathcal{L}_{adv}(x+\delta)$ 
\end{center}
where the $\mathcal{L}_{adv}$ is introduced in Eq. (\ref{eq:total-loss}), $\mathcal X$ denotes EoT~\cite{athalye2018synthesizing} distribution for robustness such as different pre-processing method.

{\bf Evaluation setup and preliminary results.} We use a similar setup in the experiment section, where we select the most practical model: SORT (Y5) and MOT17 dataset with 2080 Ti. 
The results values for R-Lat, R-Track, and \#Track are 82.3, 336.1, and 80.9, respectively, aligning closely with the findings in Table~\ref{tab:main_res}. 
These results suggest that the patch attack can potentially induce substantial system-level effects with practicality.
Note that the patch results are slightly better than the noise attack. Since the perturbation strength of the patch is much larger, it enables the attacker to generate bounding boxes with high confidence and to precisely dictate their locations. 
Thus, the tracking can be easily controlled by the attacker and the attack effects can be improved. However, the patch attack requires more complex designs such as patch size and patch location, which are correlated with the practicality and effectiveness. In this paper, we only provide a preliminary evaluation demonstrating the potential to transfer the noise attack to the patch attack. 
We leave the study on patch-based attack as our future work.

\subsection{Limitations and Future Work}
\hspace*{10pt} First, although we have some physical-world realizability improvements and some existing works demonstrate that such attack is realizable, it is still unclear whether \system can indeed work well in physical world, which can be a potential future direction to explore. Second, exploring \system under a black-box threat model, a more practical one, is another potential future work. Third,
while several availability-based latency attacks in AD have been identified, the exploration of defense directions in this context remains limited. Consequently, we consider the investigation of defenses as a part of our future work.

\section{Conclusion}
\label{sec:conclusion}
\hspace*{10pt}This paper presents a first study on availability-based latency adversarial attacks considering the entire camera-based AD perception pipeline, i.e., both object detection and object tracking.
We design a novel attack framework, \system, with a two-stage attack strategy and three novel loss functions. Our results show that \system can outperform all the existing latency attacks on camera-based object detection and significantly improve system-level effects, i.e., 95\% vehicle crash rate.
Due to the critical role of perception, we hope
that our findings and insights can inspire more future research
into this largely overlooked research perspective.

% \clearpage
\section{Acknowledgments}
We would like to thank Ziwen Wan, Yunpeng Luo, Tong Wu, Xinyang Zhang, and the anonymous reviewers for their valuable and insightful feedback. This research was supported in part by National Key R\&D Program of China (2020AAA0107702); the NSF under grants CNS-1929771, CNS-2145493, and CNS-1932464; National Natural Science Foundation of China (U21B2018, 62161160337, 62132011, 62376210, 62006181, U20B2049); Shaanxi Province Key Industry Innovation Program (2021ZDLGY01-02); and Fundamental Research Funds for the Central Universities under grant (xtr052023004, xtr022019002). Chao Shen and Qi Alfred Chen are the corresponding authors.

\bibliography{aaai24}

\end{document}